\documentclass[letterpaper, 10 pt, conference]{ieeeconf}  % Comment this line out if you need a4paper

\IEEEoverridecommandlockouts                              % This command is only needed if 
\usepackage{graphicx} % for pdf, bitmapped graphics files
\usepackage{amsmath} % assumes amsmath package installed
\usepackage{amssymb}  % assumes amsmath package installed
\usepackage[hidelinks]{hyperref}
\usepackage{subcaption}
\usepackage{comment}
\usepackage[english, ruled, linesnumbered, vlined]{algorithm2e}
\usepackage{booktabs}
\usepackage{bm}

\title{\LARGE \bf
Virtual Borders: Accurate Definition of a Mobile Robot's Workspace Using Augmented Reality
}

\author{Dennis Sprute$^{1, 2}$, Klaus T{\"o}nnies$^{2}$ and Matthias K{\"o}nig$^{1}$% <-this % stops a space
\thanks{$^{1}$The authors are with Campus Minden, Bielefeld University of Applied Sciences, 32427 Minden, Germany}%
\thanks{$^{2}$The authors are with the Faculty of Computer Science, Otto-von-Guericke University Magdeburg, 39106 Magdeburg, Germany}%
\thanks{This work is financially supported by the German Federal Ministry of Education and Research (BMBF, Funding number: 03FH006PX5).}
}

\begin{document}

\maketitle
\thispagestyle{empty}
\pagestyle{empty}

%%%%%%%%%%%%%%%%%%%%%%%%%%%%%%%%%%%%%%%%%%%%%%%%%%%%%%%%%%%%%%%%%%%%%%%%%%%%%%%%
\begin{abstract}
We address the problem of interactively controlling the workspace of a mobile robot to ensure a human-aware navigation. This is especially of relevance for non-expert users living in human-robot shared spaces, e.g. home environments, since they want to keep the control of their mobile robots, such as vacuum cleaning or companion robots. Therefore, we introduce virtual borders that are respected by a robot while performing its tasks. For this purpose, we employ a \mbox{RGB-D} Google Tango tablet as human-robot interface in combination with an augmented reality application to flexibly define virtual borders. We evaluated our system with 15 non-expert users concerning accuracy, teaching time and correctness and compared the results with other baseline methods based on visual markers and a laser pointer. The experimental results show that our method features an equally high accuracy while reducing the teaching time significantly compared to the baseline methods. This holds for different border lengths, shapes and variations in the teaching process. Finally, we demonstrated the correctness of the approach, i.e. the mobile robot changes its navigational behavior according to the user-defined virtual borders.
\end{abstract}

%%%%%%%%%%%%%%%%%%%%%%%%%%%%%%%%%%%%%%%%%%%%%%%%%%%%%%%%%%%%%%%%%%%%%%%%%%%%%%%%
\section{Introduction}
Humans and robots increasingly live together in shared spaces, such as home environments. Robots support the residents in their everyday life, e.g. as household or companion robots, and people appreciate the help of robots. But from our experience, we know that there are sometimes areas that should not be entered by a robot. These can be social places, e.g. bathrooms or bedrooms, that should be avoided by the robot due to privacy concerns. Another use case is the accurate definition of the workspace of a mobile vacuuming or mopping robot to operate in certain areas. Therefore, non-expert users need the ability to interactively and easily control a mobile robot's workspace to address this challenge.\par

For this purpose, we propose \textit{virtual borders} that are not directly visible to the user but indicate occupied areas to the robot. These are respected by the mobile robot while performing its task. We address the question of \textit{how to allow non-expert users to flexibly teach virtual borders to their robots and change their navigational behavior accordingly}. This teaching method needs to allow accurate border teaching while featuring little effort. Additionally, a feedback system giving information about learned virtual borders is desirable. We refer to a non-expert as a person that (1) has no programming skills, (2) has no experience with robotics and its insights, (3) has no cognitive impairments or upper limb disorders, but (4) has experiences with common consumer products, such as tablets or smartphones. Moreover, a non-expert (5) prefers a robust and feature-complete system to a highly sophisticated and non-intuitive one.\par

Several human-robot interfaces are imaginable for the teaching process, e.g. approaches using simple graphical user interfaces (GUIs), remote controllers, smartphones or tablets, direct physical interaction with the robot or pointing gestures (with or without auxiliary device). In order to address the above-mentioned requirements optimally, we propose a teaching method employing a RGB-D tablet, such as a Google Tango device, to interact with the robot. We choose a Tango tablet for several reasons: (1) tablet and smartphone interfaces are well established which makes them attractive for non-experts, (2) a high-accuracy on-board visual-inertial odometry allows robust 6-DoF pose tracking of the device, (3) an augmented reality~(AR) application shown on the tablet's display allows direct visual feedback to the user and (4) no additional equipment (robot, cameras or visual markers in the environment) are necessary for teaching.  
\begin{figure}
	\centering
	\includegraphics[width=0.44\textwidth]{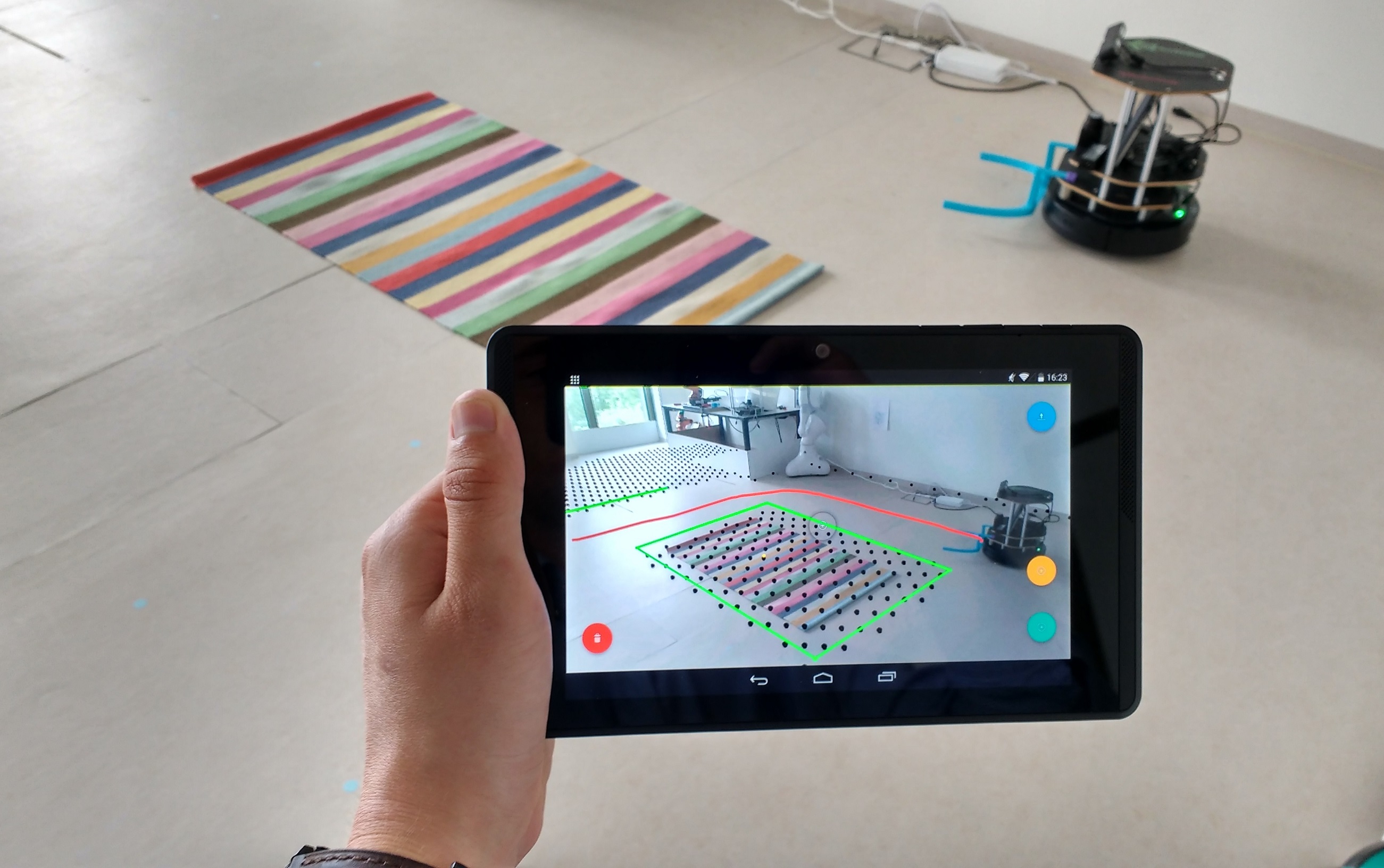}   
	\caption{A user restricts the robot's workspace using a RGB-D Google Tango tablet by specifying an area around a carpet. The robot avoids this area while working, and the tablet's AR application provides direct visual feedback to the user.}
	\label{fig:userview}
\end{figure}

Fig.~\ref{fig:userview} shows a user with a Google Tango tablet excluding a carpet area from the mobile robot's workspace. After completing the teaching process, the robot does not cross the carpet area while navigating in the environment. Our main contribution is a teaching method leveraging a RGB-D device to allow non-experts the flexible definition of a 3-DoF robot's workspace. Such a method is especially interesting for robot navigation in human-centered environments. To the best of our knowledge, this is the first time a mobile \mbox{RGB-D} tablet or smartphone is used for human-robot interaction.\par

The remainder of this paper is structured as follows: in the next section, we give an overview of related work concerning the topic before we formally define the problem. Subsequently, we give details about our proposed teaching method based on a RGB-D device. We also evaluate the proposed method concerning accuracy, teaching time and correctness and compare the results with selected baseline methods. These experimental results are presented in the following section. Finally, we discuss our method concerning strengths and weaknesses and point out work for the future.

\section{Related Work}
There are different types of maps that differ in the way they model the environment, e.g. metric maps represent geometric properties of the environment. A typical representation of this category is an occupancy grid map~(OGM)~\cite{Moravec:1985} that is widely used in robot navigation and path planning. It models the environment by means of cells containing a probability for the occupancy of the corresponding area. In order to create an OGM of an environment and localize the robot with respect to it, Simultaneous Localization and Mapping (SLAM) algorithms~\cite{Thrun:2005} are widely established. Cadena et al. give a comprehensive overview of the evolution of SLAM from the past to the future~\cite{Cadena:2016}.\par

Along the occupancy information modeled in an OGM, maps can contribute additional information, such as semantics~\cite{Kostavelis:2015} or social information~\cite{Kruse:2013}. Especially social information can be used with the purpose of changing the robot's navigational behavior in human-centered environments, e.g. O'Callaghan et al. incorporate motion patterns of people into the robot's trajectories~\cite{O'Callaghan:2011}, and Alempijevic et al. jointly learn a map from robots' sensor measurements and human trajectories as basis for path planning~\cite{Alempijevic:2013}. Other works use social costmaps built from sensor measurements to realize a human-aware navigation~\cite{Talebpour:2016}, integrate social norms into the costmap to change the way a robot approaches a human~\cite{Ramirez:2016} and propose human motion maps to represent the distribution of human motion in a map~\cite{Ogawa:2014}. A survey on recent trends in socially aware robot navigation and a historical overview is given by Charalampous~et~al.~\cite{Charalampous:2017}.\par

These implicit approaches to change the robot's navigational behavior are based on observations. They are user-friendly because no explicit interaction is necessary, but they are not flexible enough for the problem addressed in this work. We argue that the teaching of \textit{arbitrary} virtual borders can only be accomplished through explicit user interaction, e.g. drawing boundaries on a previously created OGM that today's home robots already provide. However, this solution is not suitable to define \textit{accurate} virtual borders since it is hard to correspond points in the OGM with points in the (featureless) environment. Other examples for this category are a GUI-based user interface to sketch the area for a vacuum cleaning robot~\cite{Sakamoto:2009} and a virtual wall system based on beacon devices~\cite{Chiu:2011}. The first approach needs several top-view cameras installed in the environment to stream images to the user's display, while the latter is restricted by the conic beam of the beacon devices. These only allow the blocking of certain areas using a straight line while consuming power and being intrusive. Magnetic stripes placed on the ground known from commercial vacuum cleaning robots are intrusive as well. To address these aspects of intrusiveness, power-consumption and small flexibility, Sprute et al. propose a framework for interactive teaching of virtual borders and an implementation based on visual markers~\cite{Sprute:2017} and a laser pointer~\cite{Sprute:2017b}. Although both approaches are flexible and allow teaching of arbitrary virtual borders, they do not provide an inherent feedback system and their teaching time is linear with respect to the border length.\par

To purposely address this lack and for the reasons mentioned in the introductory part, we chose a RGB-D Google Tango device as interaction device. It has been used in several robotics-related applications, e.g. indoor-localization given a 2D floor plan~\cite{Winterhalter:2015} and real-time 3D reconstruction~\cite{Klingensmith:2015}. Other use cases include optimization of SLAM by text spotting~\cite{Wang:2015} or controlling of a quadrotor equipped with a Tango smarthpone~\cite{Loianno:2015}. These applications show the potential of mobile RGB-D devices in the context of robotics.

\section{Problem Statement}
Before we give details on the proposed teaching method, we introduce the notation we use throughout the paper and formally define the problem of interactively manipulating an OGM using a RGB-D device. It is the goal to change the robot's navigational behavior in future tasks according to the users' needs. An OGM models the physical environment in terms of cells containing probabilities for the occupancy of the corresponding area. $M(x, y) \in [0, 1]$ denotes the occupancy probability for the cell $(x, y)$ in the map $M$. Furthermore, we define all possible coordinates $(x,y) \in \mathbb{R}^2$ of a map $M$ as the domain of the map  $\Omega(M) \subset \mathbb{R}^2$. At the beginning, an OGM of the physical environment $M_{prior}$ containing walls and furniture is given. Due to the iterative nature of the teaching method, $M_{prior}$ can also contain virtual borders from previous teaching processes. Since we want to integrate virtual borders into the map, the user defines a manipulation so that $M_{prior} \rightarrow M_{posterior}$. This posterior map $M_{posterior}$ contains the physical environment as well as the user-defined virtual borders and can be used for socially aware navigation and path planning.

\section{Teaching Using a RGB-D Device}
We propose a teaching method based on a RGB-D tablet in combination with an AR application to address the problem of interactive teaching of virtual borders and changing the mobile robot's navigational behavior accordingly. To this end, a person uses a Google Tango tablet to move around in the environment and to select points on the ground plane by interacting with the mobile device. The tablet simultaneously acts as a feedback device showing an augmented live video of its on-board camera. The user-defined virtual borders are integrated into the prior map of the environment to ensure a human-aware navigation. A full video of a teaching process can be found in the supplementary video at: \url{https://youtu.be/oQO8sQ0JBRY}.

\subsection{Requirements}
\label{sec:requirements}
\begin{figure}
	\centering
	\includegraphics[width=0.49\textwidth]{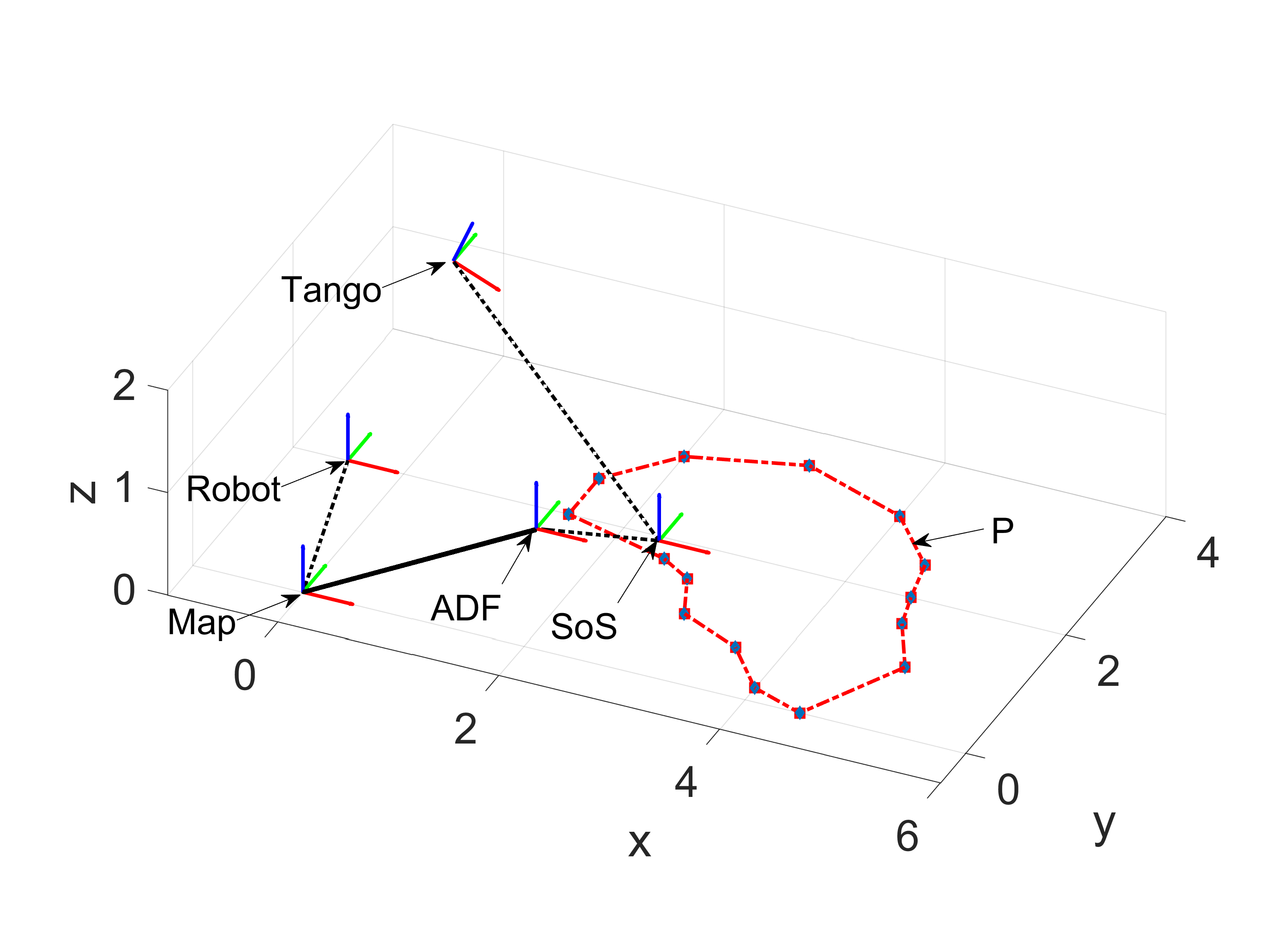}   
	\caption{Relevant coordinate frames and their relations shown as black lines. The solid black line between the \textit{ADF} and \textit{Map} coordinate frames indicates the manual registration. The red line depicts an illustrative virtual border polygon $\mathcal{P}$.}
	\label{fig:coordinateFrames}
\end{figure}

In order to realize this behavior, the robot as well as the Tango tablet need to have access to the same global OGM $M_{prior}$ that was created in advance. For this purpose, we relate the relevant coordinate frames to each other as shown in Fig.~\ref{fig:coordinateFrames}. We assume the Tango device to be localized within the environment. The origin of this previously constructed environment is the \textit{ADF} (Area Description File) coordinate frame. When starting the teaching application, the \textit{SoS}~(Start~of~Service) coordinate frame marks the current pose of the Tango device. While localizing in the environment employing visual features, the transformation between \textit{SoS} and \textit{ADF} is established. The Tango device uses its accurate on-board visual-inertial odometry to keep track of its current pose \textit{Tango} with respect to \textit{SoS}. The \textit{ADF} coordinate frame is manually related to the \textit{Map} coordinate frame once a visual model of the environment is learned. Finally, the dynamic pose of the mobile robot \textit{Robot} is related to the \textit{Map} frame using adaptive Monte Carlo localization~(AMCL)~\cite{Fox:2003} using the robot's laser scanner. This ensures transformations between all relevant coordinate frames. All of these transformations belong to $SE(3)$.

\subsection{Area Definition}
We define a virtual border as a triple $V = (\mathcal{P}, \bm{s}, \delta)$ where each component is specified in the interactive teaching process using the RGB-D device. It can perceive a 3D point cloud of its environment that is used to specify $n$ points $\bm{P}_i \in \mathbb{R}^3, 1 < i < n$ on the ground plane. Only points on the ground plane are of interest because the mobile robot operates in the plane. By transforming these points into the \textit{Map} coordinate frame, we obtain $n$ points $\bm{p}_i \in \mathbb{R}^2$ building a polygonal chain $\mathcal{P}$:
\begin{equation}
\mathcal{P} = \bigcup_{i=1}^{n-1} [\bm{p}_i \bm{p}_{i+1}],
\end{equation}
with 
\begin{equation}
[\bm{p}_i \bm{p}_{i+1}] = \{(1-\lambda)\bm{p}_i + \lambda \bm{p}_{i+1} \mid \lambda \in [0,1]\}
\end{equation}
being a line segment between two points. We distinguish between simple and closed polygonal chains to define arbitrary areas in the environment as polygons or separating curves.\par

Additionally, the user employs the RGB-D device to select a seed point $\bm{s} \in \mathbb{R}^3$ that indicates the area to be manipulated. The corresponding cell in the global OGM is denoted as $\bm{s^*} \in~\Omega(M_{prior})$. Finally, the user has the possibility to specify the occupancy probability $\delta \in [0, 1]$ for the area indicated by $\bm{s}$. 

\subsection{Map Creation}
After defining a virtual border $V$, we use the polygonal chain $\mathcal{P}$ to partition the map into two areas:
\begin{equation}
A_c = \{\bm{c} \in \Omega(M_{prior}) \mid \bm{c}\ connected\ to\ \bm{s}^* \},
\end{equation}
which is the area that is directly connected to the cell corresponding to the seed point $\bm{s}^*$ and 
\begin{equation}
A_{nc} = \Omega(M_{prior}) \setminus A_c,
\end{equation}
which is the complementary area containing coordinates disconnected from the seed point $\bm{s}^*$. Two cells $\bm{a} \in \Omega(M)$ and $\bm{b} \in \Omega(M)$ in a map $M$ are $connected$ if:
\begin{equation}
\begin{split}
\exists f: [0..1] \rightarrow \Omega(M): &f(0)=\bm{a}, f(1)=\bm{b},\\
&\forall i, j \in [0..1]: M(f(i)) = M(f(j))
\end{split}
\end{equation}
where $f$ is a continuous mapping.\par

If the border polygon $\mathcal{P}$ is a simple polygonal chain, we linearize the first $[\bm{p_1 p_2}]$ and the last $[\bm{p_{n-1} p_n}]$ line segments to partition the map. An example for such a simple polygonal chain and its linearization is shown in Fig.~\ref{fig:teach2} where the green line indicates the actual polygonal chain $\mathcal{P}$ and the black dots the resulting occupied space. The system automatically extends the virtual border to the borders of the prior map $M_{prior}$. This allows the user to easily exclude large areas from the robot's workspace with a single curve or line. Finally, we construct the posterior map $M_{posterior}$ dependent on the given prior map $M_{prior}$ and the components of the virtual border $V$ as follows:
\begin{equation}
M_{posterior}(x, y)=
\begin{cases}
	\delta& if\ (x, y) \in A_c\\
	M_{prior}(x, y)& if\ (x, y) \in A_{nc}
\end{cases}
\end{equation}

By iterating this teaching process $N$ times and defining a sequence of virtual borders $V^{*}=\{V_1, V_2, ..., V_N\}$, the user can define arbitrary virtual borders in the environment. This allows the flexible definition of a mobile robot's workspace.

\subsection{Interaction \& Feedback}
\begin{figure*}
		\centering
        \begin{subfigure}[b]{0.24\textwidth}
                \centering
                \includegraphics[width=\textwidth]{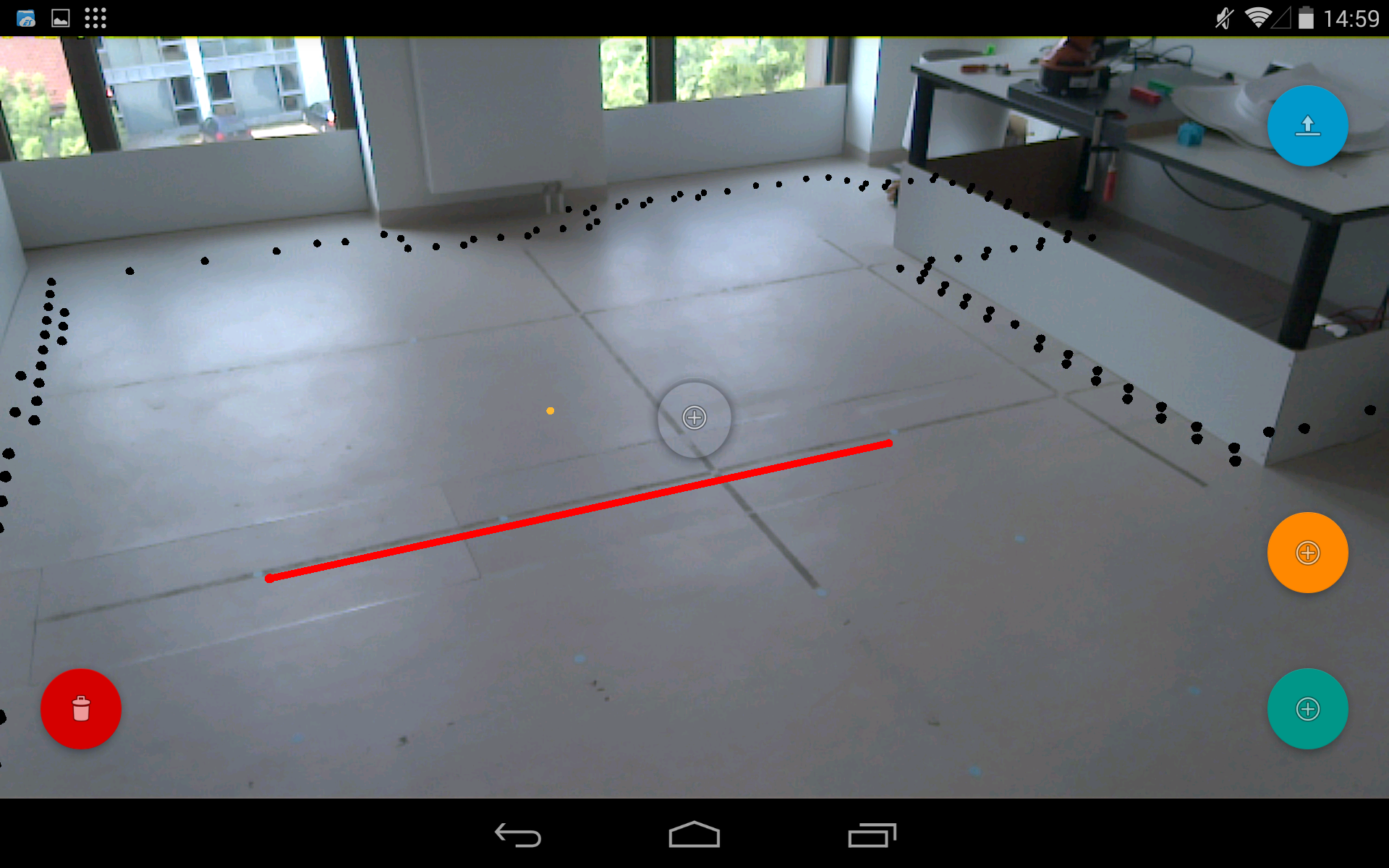}
                \caption{}    
                 \label{fig:teach1}                        
        \end{subfigure}        
        \centering
        \begin{subfigure}[b]{0.24\textwidth}
                \centering
                \includegraphics[width=\textwidth]{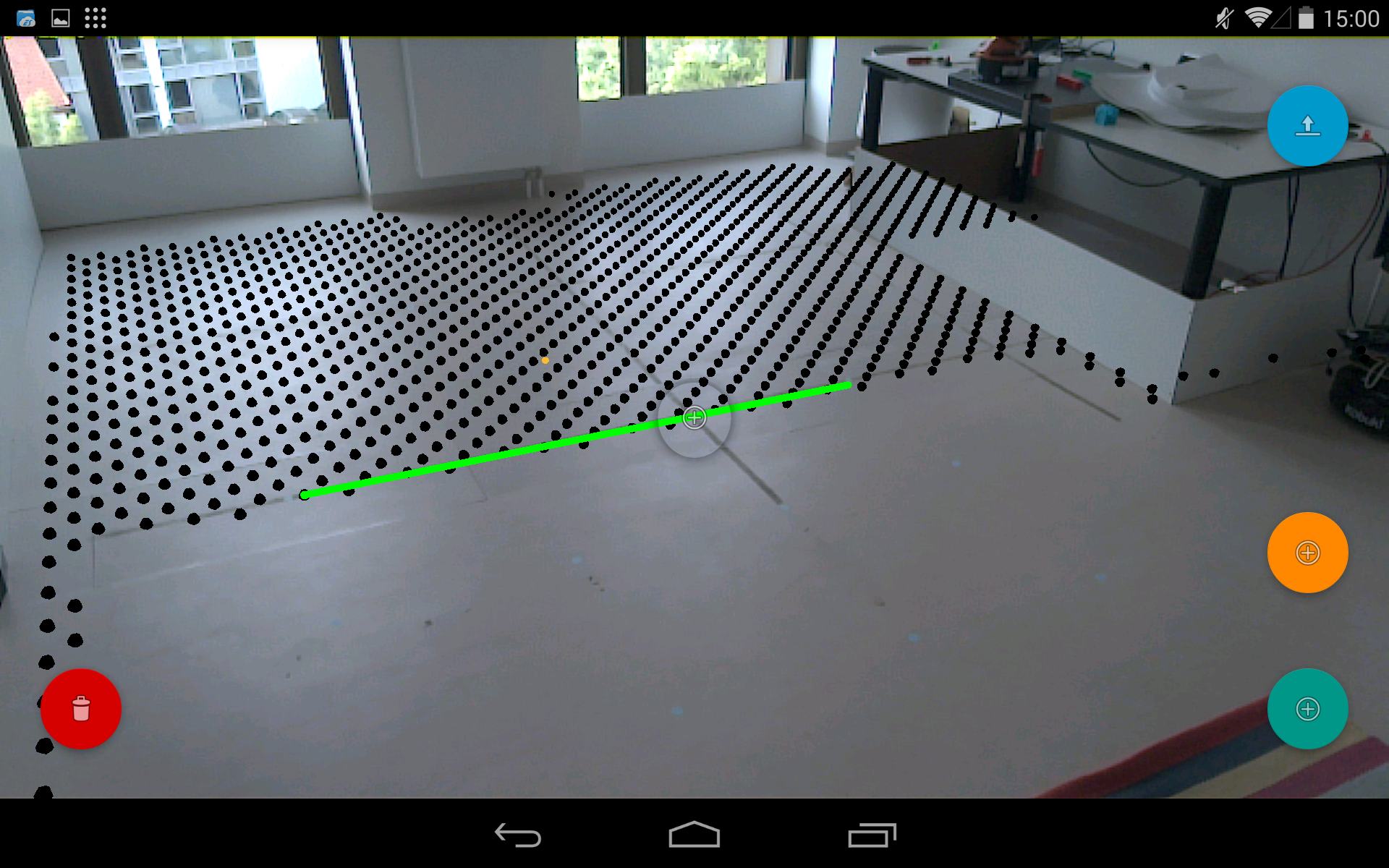}
                \caption{}       
                \label{fig:teach2} 
        \end{subfigure}          
        \centering
        \begin{subfigure}[b]{0.24\textwidth}
                \centering
                \includegraphics[width=\textwidth]{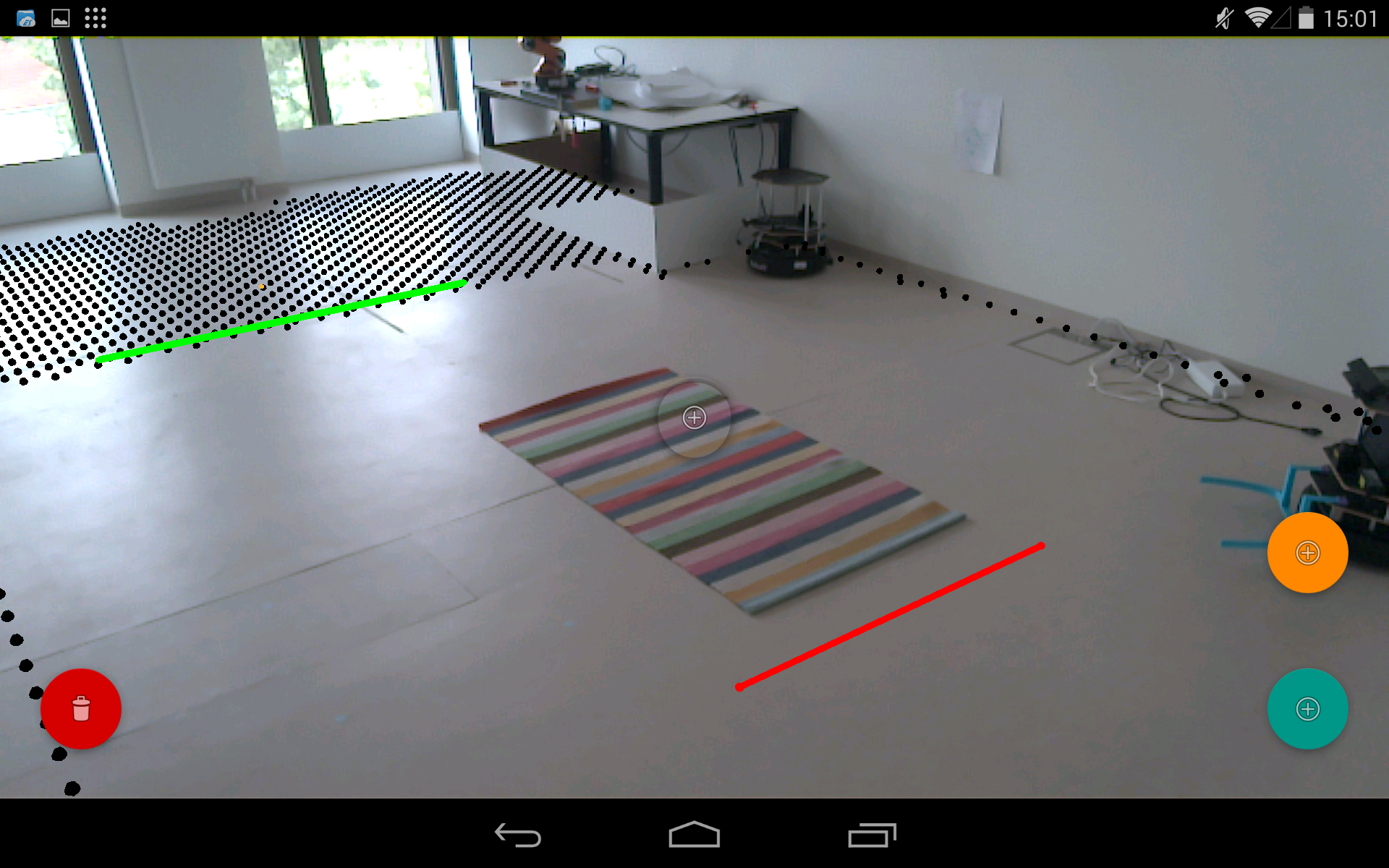}
                \caption{}     
                 \label{fig:teach3}            
        \end{subfigure}  
        \centering
        \begin{subfigure}[b]{0.24\textwidth}
                \centering
                \includegraphics[width=\textwidth]{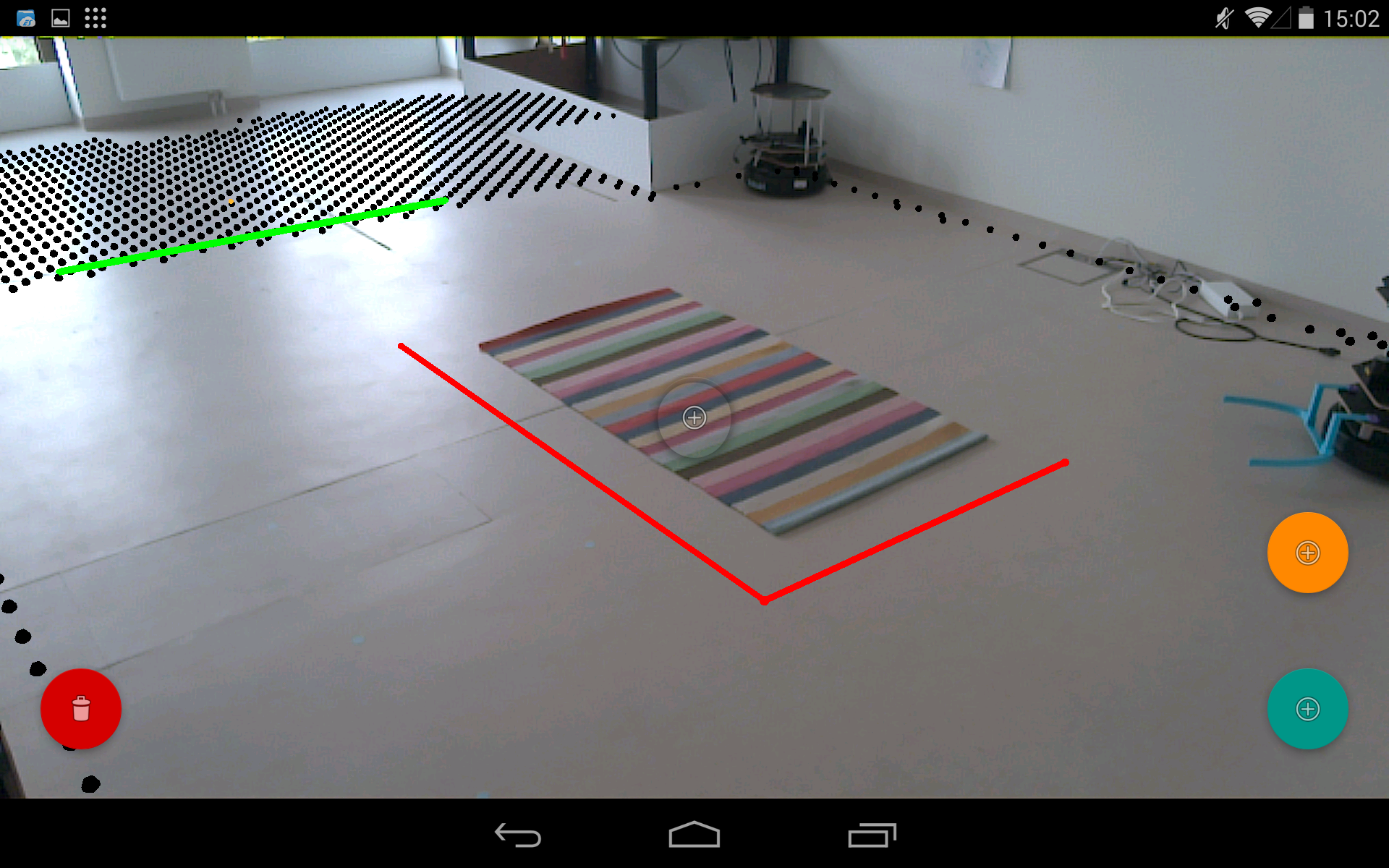}
                \caption{} 
                 \label{fig:teach4}                
        \end{subfigure}   
        \centering
        \begin{subfigure}[b]{0.24\textwidth}
                \centering
                \includegraphics[width=\textwidth]{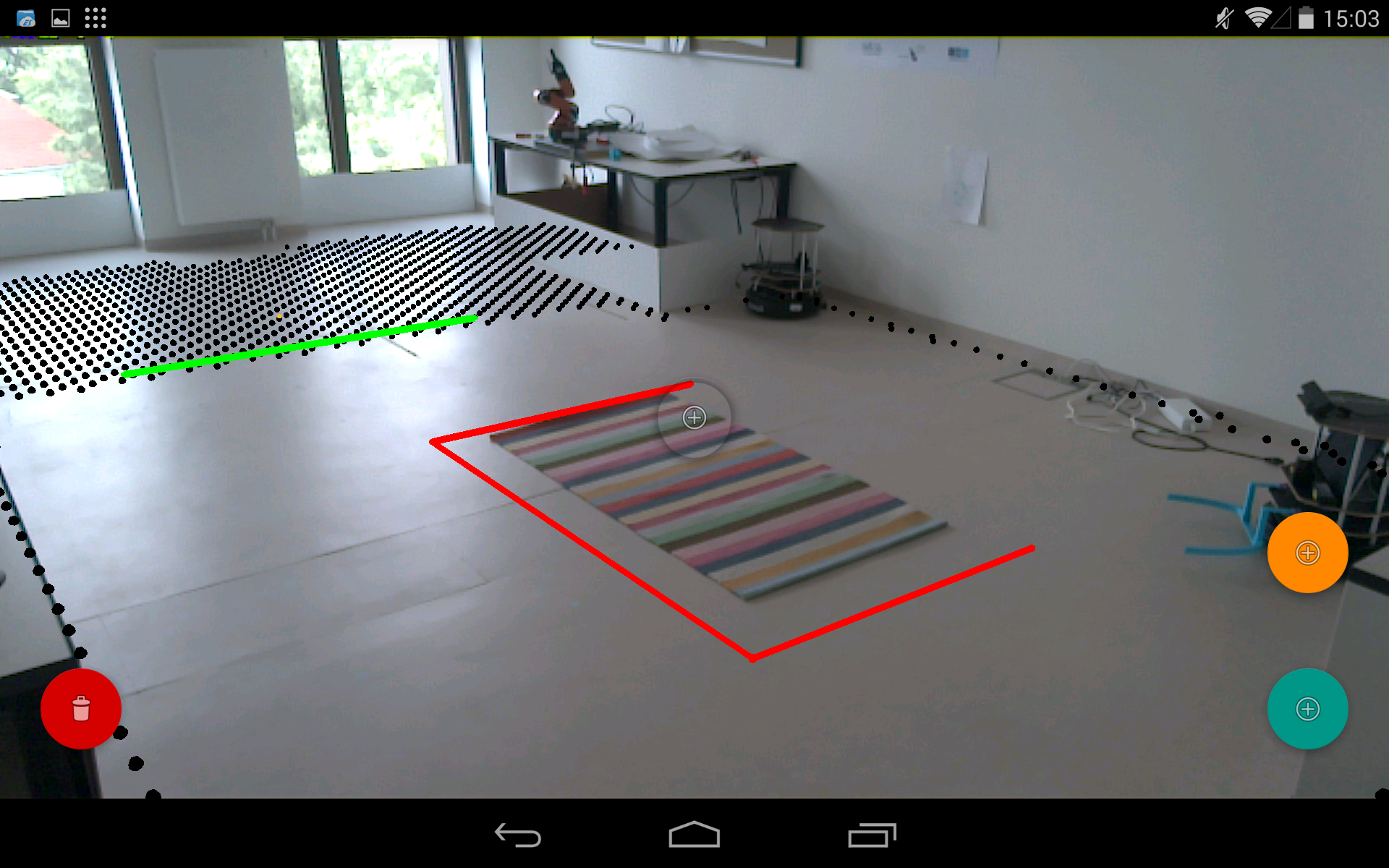}
                \caption{}     
                 \label{fig:teach5}            
        \end{subfigure}        
        \centering
        \begin{subfigure}[b]{0.24\textwidth}
                \centering
                \includegraphics[width=\textwidth]{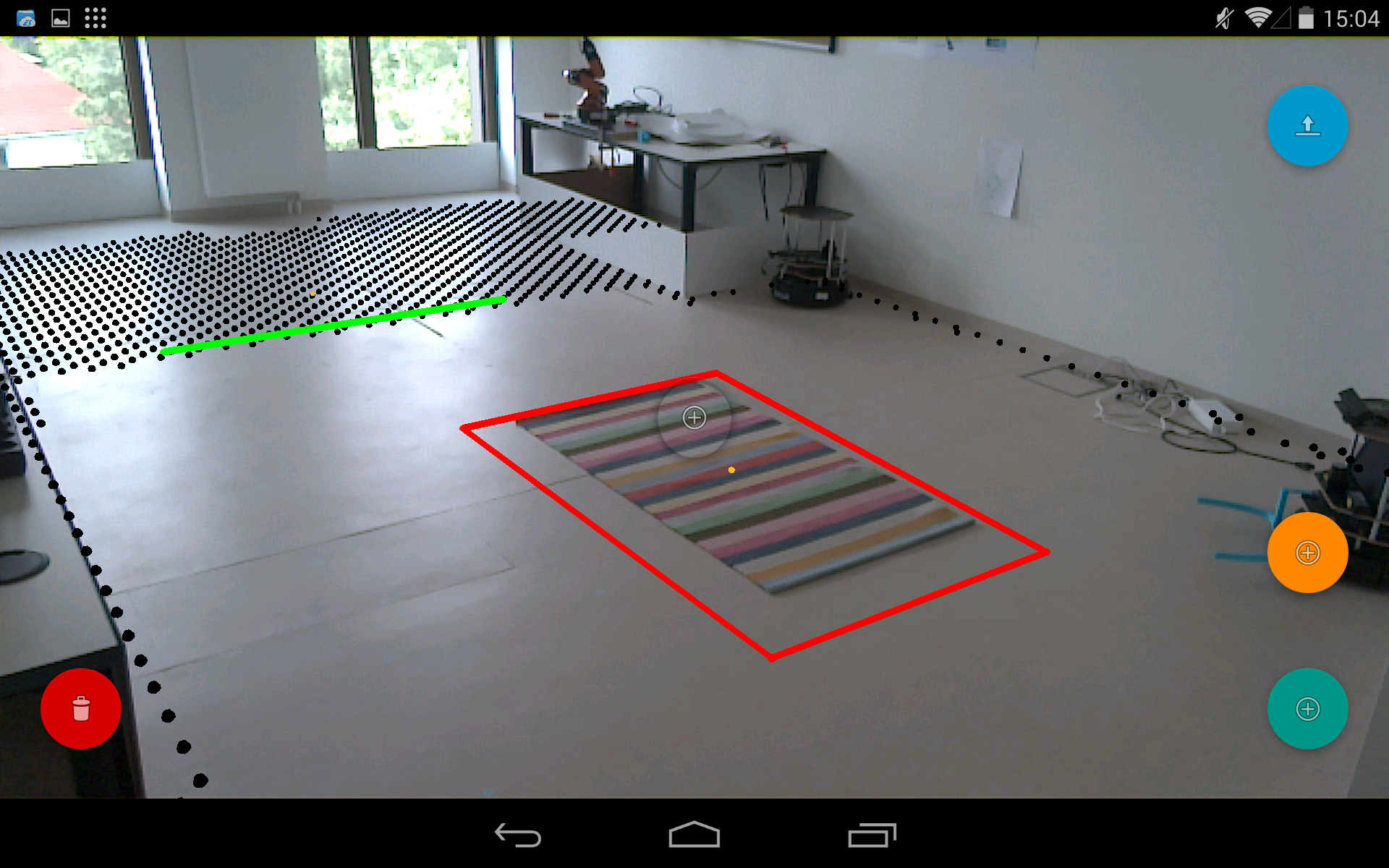}
                \caption{}     
                 \label{fig:teach6}            
        \end{subfigure}          
        \centering
        \begin{subfigure}[b]{0.24\textwidth}
                \centering
                \includegraphics[width=\textwidth]{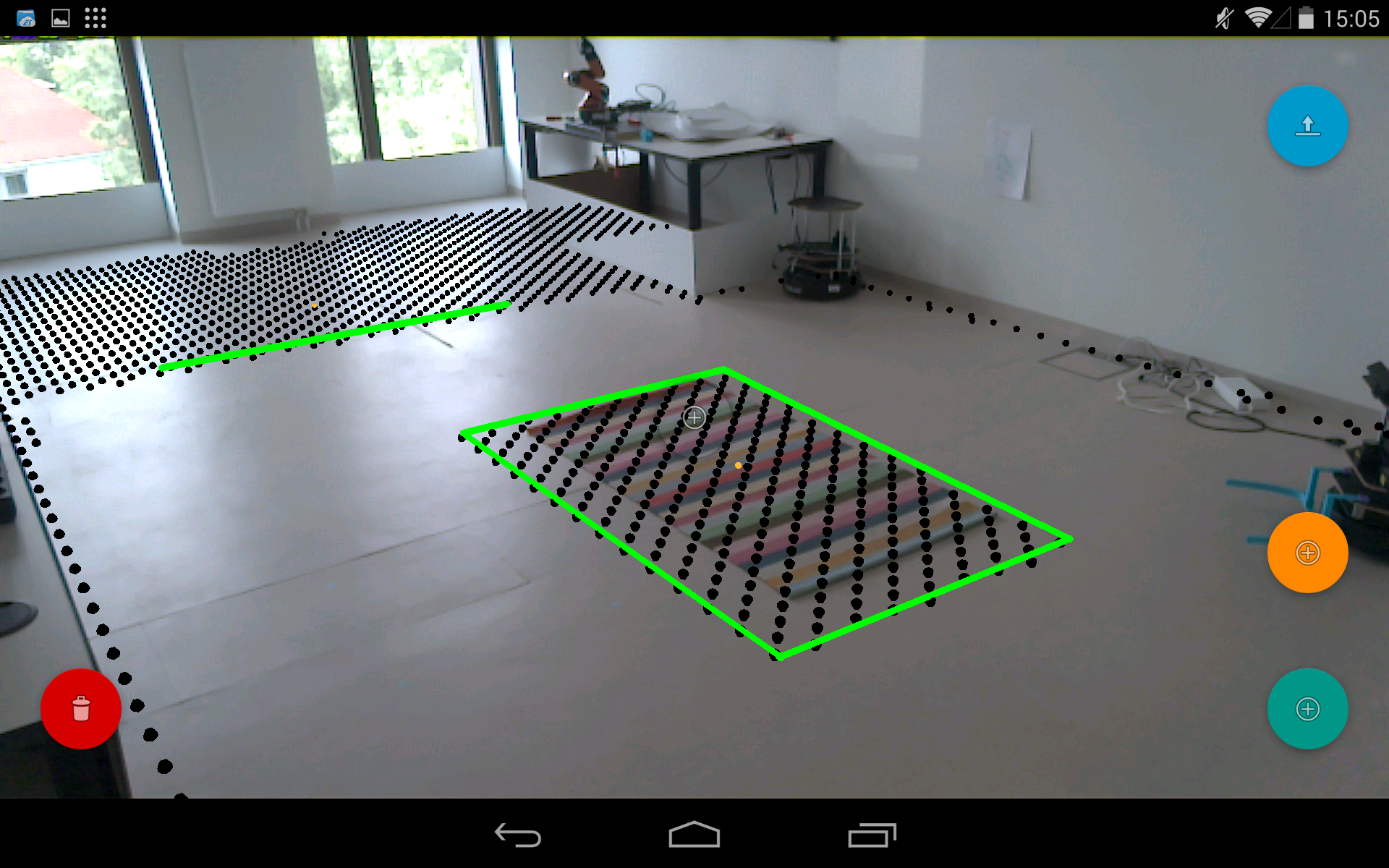}
                \caption{}     
                 \label{fig:teach7}            
        \end{subfigure}  
        \centering
        \begin{subfigure}[b]{0.24\textwidth}
                \centering
                \includegraphics[width=\textwidth]{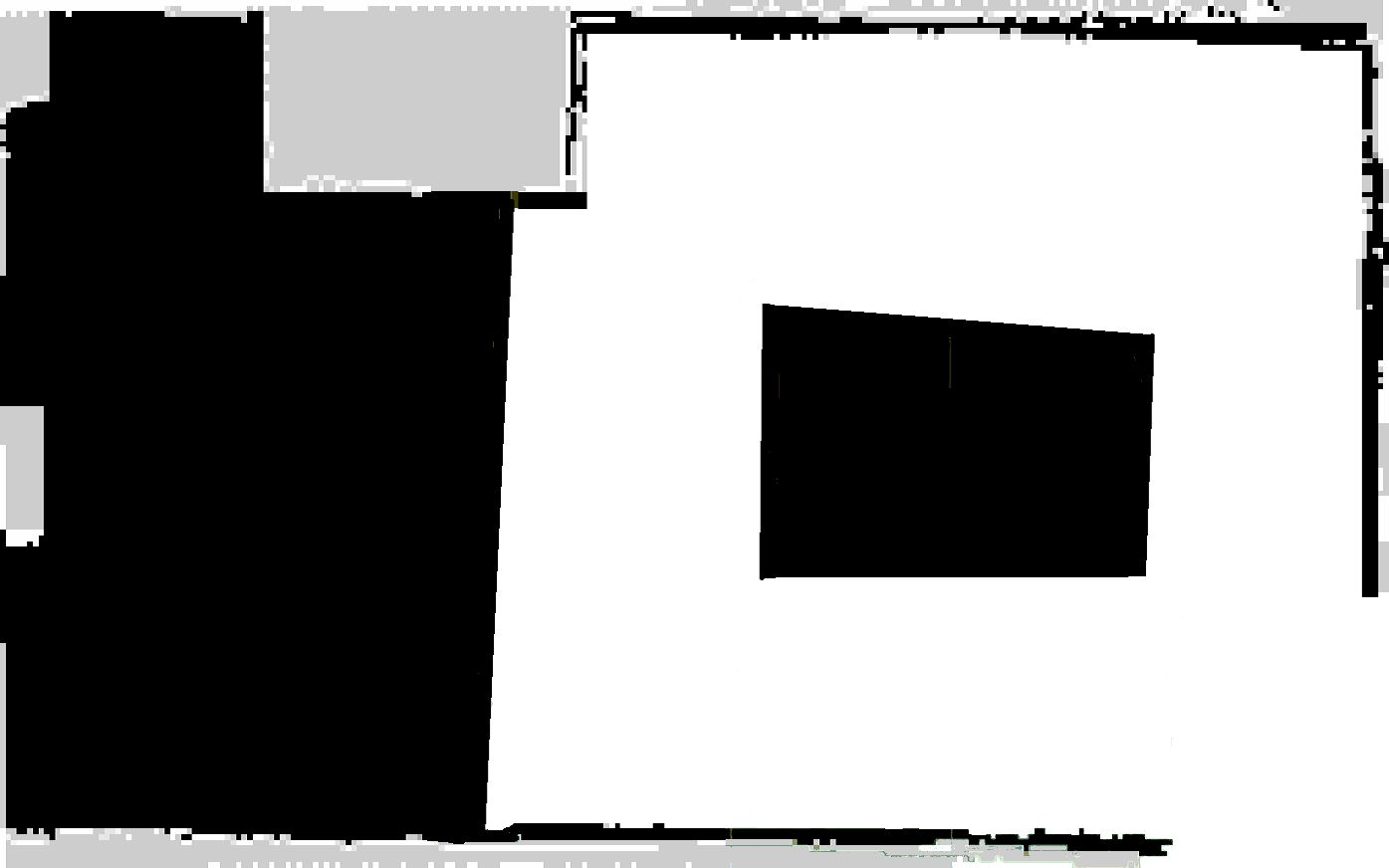}
                \caption{}    
                 \label{fig:teach8}             
        \end{subfigure}    
        \caption{Screenshots of the teaching process. Red lines show boundaries of an area drawn by the user, while green lines indicate the user that the lines have been successfully integrated into the prior OGM. Black dots visualize the occupied areas in the global OGM. (a) The user defines a first border as a separating curve to exclude the window area  from the workspace, and (b) shows the area integrated into the global OGM. (c) - (f) The user defines a second border as a polygon around the carpet to avoid the robot from crossing it. (g) - (h) show the final maps from the user's and mobile robot's perspective.} 
        \label{fig:teaching}
\end{figure*}
The user only needs the Google Tango tablet with its AR application to specify a virtual border $V$ with all its components. The person moves around in the environment with the tablet and selects virtual border points $\mathcal{P}$ by pointing the device towards the desired points on the ground plane. The seed point $\bm{s}$ is selected analogously, and a simple menu allows the definition of the occupancy probability $\delta$. Simultaneously, the Tango's camera image augmented with the virtual border points $\mathcal{P}$ is displayed on the tablet's screen. Additionally, the global OGM, that is the basis for navigational costmaps, is integrated into the view. This makes it easy for the user to understand the workspace of the mobile robot. Besides, the user immediately gets visual feedback by the system and can correct eventual mistakes. This is a crucial feature that has not been addressed in previous works yet. Fig.~\ref{fig:teaching} shows some screenshots of the teaching process where the AR application on the Tango device is used to specify two virtual borders: a separating curve and a polygon. 

\section{Evaluation}
We provide quantitative and qualitative results for our proposed method concerning three criteria: accuracy, teaching time and correctness. The results are compared with two other teaching methods that are described in the following subsection. For the proposed method, we used a Google Tango tablet as mobile RGB-D device to acquire depth measurements and colored images from the environment. All methods were implemented as a ROS package, and we performed the following experiments on OGMs with a resolution of 2.5 cm per pixel in our 6.1~m $\times$ 3.5~m lab environment. A prior map $M_{prior}$ of the environment was created with a common SLAM algorithm~\cite{Grisetti:2007} using a particle filter and the robot's on-board laser scanner to obtain measurements.

\subsection{Baseline Methods}
The baseline methods allow the incorporation of virtual borders into a given prior map, but do not consider other occupancy probabilities except of \textit{free} and \textit{occupied}. Despite this limitation, both methods are the most flexible ones mentioned in the literature and are suitable for the evaluation of the accuracy and the teaching time. Since both comparative methods require a robotic platform for teaching, their evaluation is based on a TurtleBot v2 equipped with a laser scanner and a front-mounted \mbox{RGB-D} camera.

\subsubsection{Marker~\cite{Sprute:2017}}
The first method employs visual markers to teach virtual borders to a mobile robot. The user guides the mobile robot by showing visual markers, and the robot records its trajectory while following the marker. Different marker IDs indicate different states of the teaching process, e.g. recording borders or defining a seed point. The trajectory is used to define the virtual border in the environment.

\subsubsection{Pointer~\cite{Sprute:2017b}}
This method uses a laser pointer as human-robot interface and allows the user to define arbitrary areas in an environment. The user guides the robot with a laser spot on the ground, and the robot employs its trajectory to define the virtual border. If the laser point leaves the mobile robot's field of view, the robot follows the direction of the laser point. Visual Morse code is used to switch between different states of the teaching process like marker IDs in the previously described method.

\subsection{Experimental Procedure}
We conducted two different experiments to evaluate the proposed teaching method and compared it with the baseline methods. The split of the experiments was necessary because it was not possible to evaluate all teaching methods on a large dataset thoroughly with multiple users. It would have taken several hours per participant. Therefore, we conducted two different experiments that complemented each other as explained below.

\subsubsection{Experiment 1}
This experiment was performed by 15 non-experts (9 male, 6 female) who rated their robotic skills on an 11-point Likert scale (0 - 10) less than 5. Their ages ranged between 16 years and 56 years with a mean age of 34.87 years and standard deviation of 13.42 years. For the purpose of this experiment, we placed a common carpet (2.00~m $\times$ 1.25~m) on a fixed position on the ground similar to Fig.~\ref{fig:lab} and manually created a ground truth map for this scenario. This should reflect a typical use case of the proposed system. Before conducting the experiment, each participant had some time to get familiar with the interaction devices and teaching methods, i.e. guiding the mobile robot or using the AR application. Afterwards, a participant was asked to exclude the carpet area from the workspace of the mobile robot. An experimenter documented the resulting posterior maps and teaching times for each participant and interaction device. The duration started with the selection of the first border point and ended on completion of the posterior map. Each participant conducted this experiment with all three interaction devices (\textit{within-subject design}). The order of the interaction devices was randomized to avoid order effects. The advantage of this experiment is the participation of multiple users in the evaluation, but their performance was limited to a single map for each interaction~device.

\subsubsection{Experiment 2}
In order to consider different maps in the teaching process, we also evaluated our method on a self-recorded dataset containing ten different maps with different polygonal-shaped virtual borders that were manually integrated into the OGM of the lab environment beforehand. The lengths of the virtual borders ranged from 4~m to 13~m increasing by 1~m (map 1: 4~m, map 2: 5~m, ..., map 10: 13~m), and their shapes were convex and non-convex. Three example ground truth maps of the dataset are visualized in the first column of Fig.~\ref{fig:accVisualization}. Due to practical reasons, all runs of this experiment were performed by a single non-expert user (male, 26 years). Since he performed all runs (in total 150 runs), he represented an experienced non-expert who got familiar with the interaction devices and teaching methods. The non-expert was asked to specify virtual borders according to the ground truth maps in the dataset using one of the interaction devices. This ground truth data was conveyed to the user in form of small markers on the ground. We performed five runs for each map to introduce some variation into the teaching process resulting in 50 runs per interaction device. Different start positions of the mobile robot or the Tango tablet were considered as variations. Similar to \textit{Experiment~1}, the teaching time and resulting posterior maps were documented. The results of the baseline methods were taken from ~\cite{Sprute:2017}, \cite{Sprute:2017b}. The strengths of this experiment are the consideration of different border lengths, shapes and variations in the teaching process. The drawback is the evaluation with only one non-expert.

\subsection{Accuracy}
\begin{figure*}
	\centering
	\includegraphics[width=\textwidth]{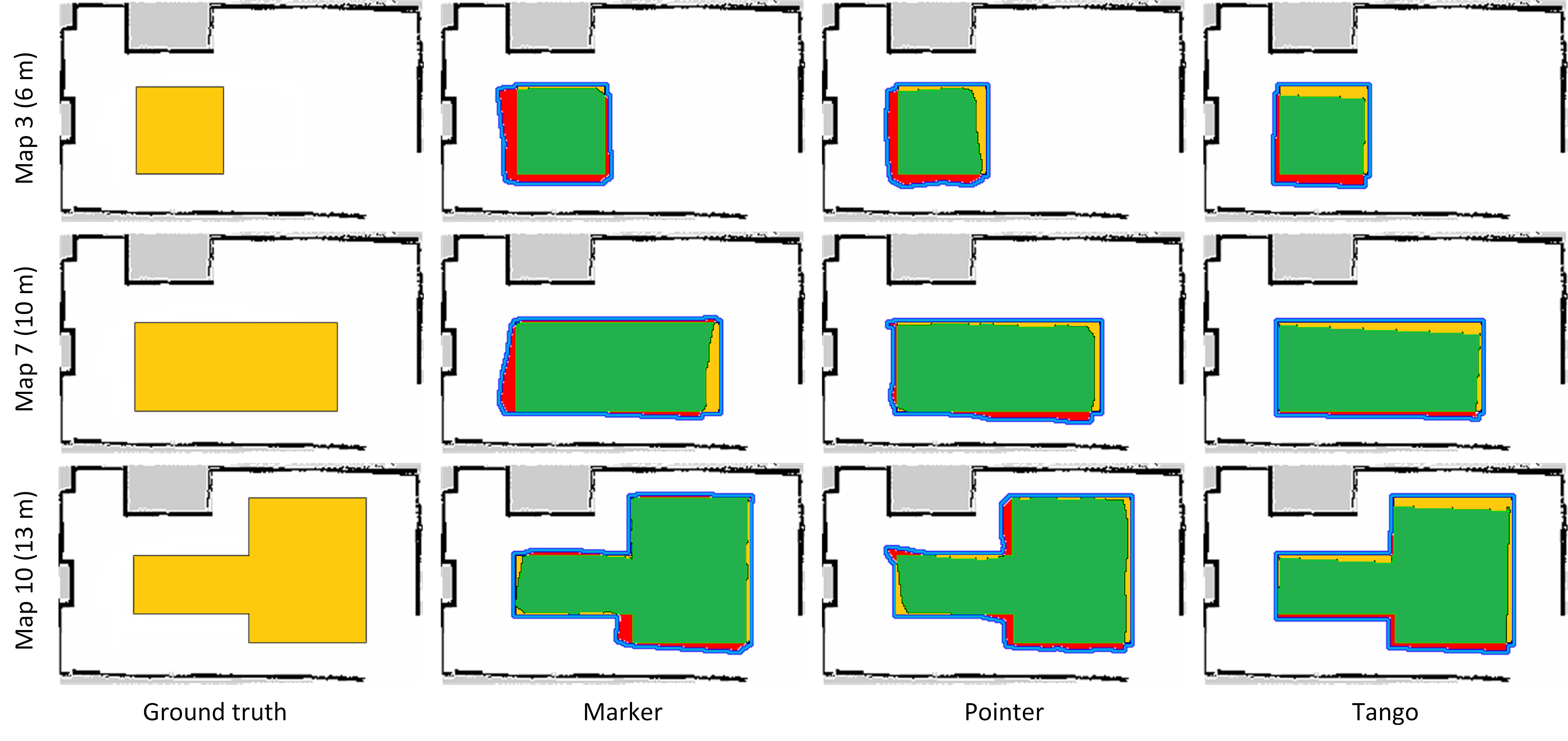}   
	\caption{Qualitative accuracy results for different virtual border maps and interaction devices. The first column shows three different ground truth virtual border maps taken from the self-recorded dataset. The other columns show the overlapping ground truth and user-defined maps with respect to the teaching methods. Colors are only used for visualization purposes.}
	\label{fig:accVisualization}
\end{figure*}

The evaluation of this criterion answers the question of \textit{how accurate are the virtual borders transferred from a user to the system}. Accurately user-defined borders are especially important for a task such as vacuum cleaning around a carpet. In order to assess the accuracy of a virtual border specified by the user, we consider the Jaccard index between two virtual areas $GT$ and $UD$ as similarity score:
\begin{equation}
J(GT, UD) = \dfrac{|GT \cap UD|}{|GT \cup UD|} \in [0, 1]
\end{equation}
These two variables are defined as follows:
\begin{enumerate}
	\item GT (ground truth): This set contains all cells of the OGM that belong to the ground truth virtual area that was manually created before evaluation. It is visualized as yellow pixels in the first column and yellow and green pixels in the remaining columns of Fig~\ref{fig:accVisualization}.
	\item UD (user defined): This set contains all cells of the OGM that belong to a user-defined virtual area that was defined by the user in the teaching process. It is visualized as red and green pixels in the last three columns of Fig~\ref{fig:accVisualization}.
\end{enumerate}
$|GT \cap UD|$ is the number of overlapping pixels between the ground truth and user-defined areas, whereas $|GT \cup UD|$ is the size of the union set. The Jaccard index can be visually interpreted as the size of the green area with respect to the area enclosed by the blue contour in Fig~\ref{fig:accVisualization}. Since this measure is independent of the size of the map, it can be used to compare different teaching methods easily.

\begin{figure}
	\centering
	\includegraphics[width=0.45\textwidth]{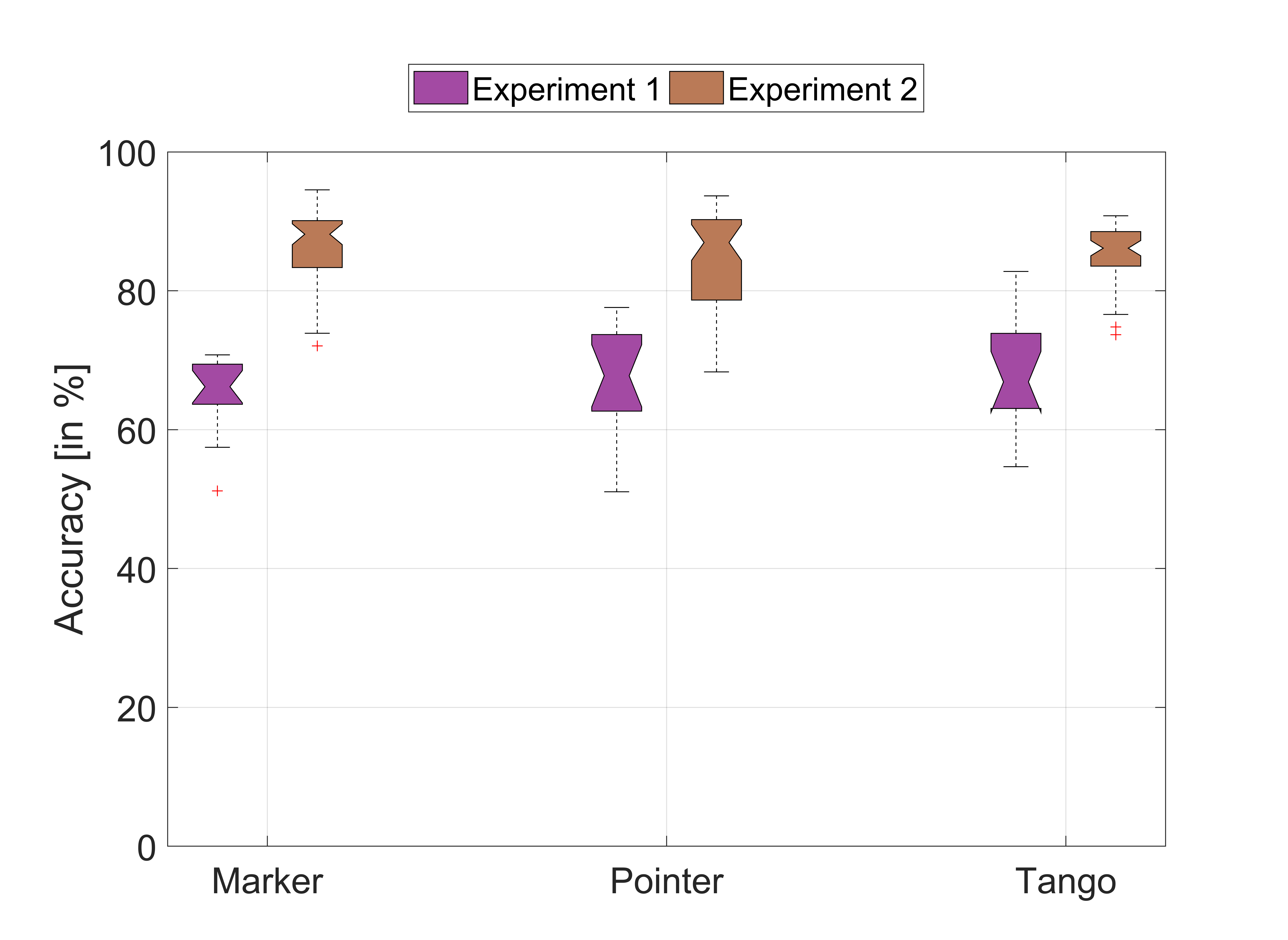}   
	\caption{Comparison of quantitative accuracy results for both experiments depending on the interaction device.}
	\label{fig:accComparison}
\end{figure}
The accuracy results for \textit{Experiment~1} are shown as purple box plots in Fig.~\ref{fig:accComparison}. The results range from a mean accuracy of 65.5\% for the marker approach to a mean accuracy of 68.5\% for our proposed approach. The laser pointer method yields a mean accuracy of 66.4\% and is placed between the other methods. Overall, there is no significant difference between the interaction devices. Besides, the standard deviations range from 5.2\% for the marker to 8.5\% for the Tango approach. The brown box plots depict the accuracy results for \textit{Experiment~2} dependent on the interaction devices. Since this is a single-user experiment, the values are taken from all runs performed in the experiment. It is apparent that the mean accuracies for \textit{Experiment~2} are better (marker: +21.4\%, pointer: +18.2\%, Tango: +16.8\%). This is because \textit{Experiment~2} was performed by a single non-expert who gained some experience during the teaching process, while \textit{Experiment~1} was performed by non-experts who defined virtual borders for the first time. Thus, experience in using the interaction devices and teaching methods can increase the accuracy. Despite this difference, the accuracies of all interaction devices also feature an equally high accuracy in \textit{Experiment~2} (marker: 86.6\%, pointer: 84.6\%, Tango:~85.3\%).\par

\begin{figure}
	\centering
	\includegraphics[width=0.45\textwidth]{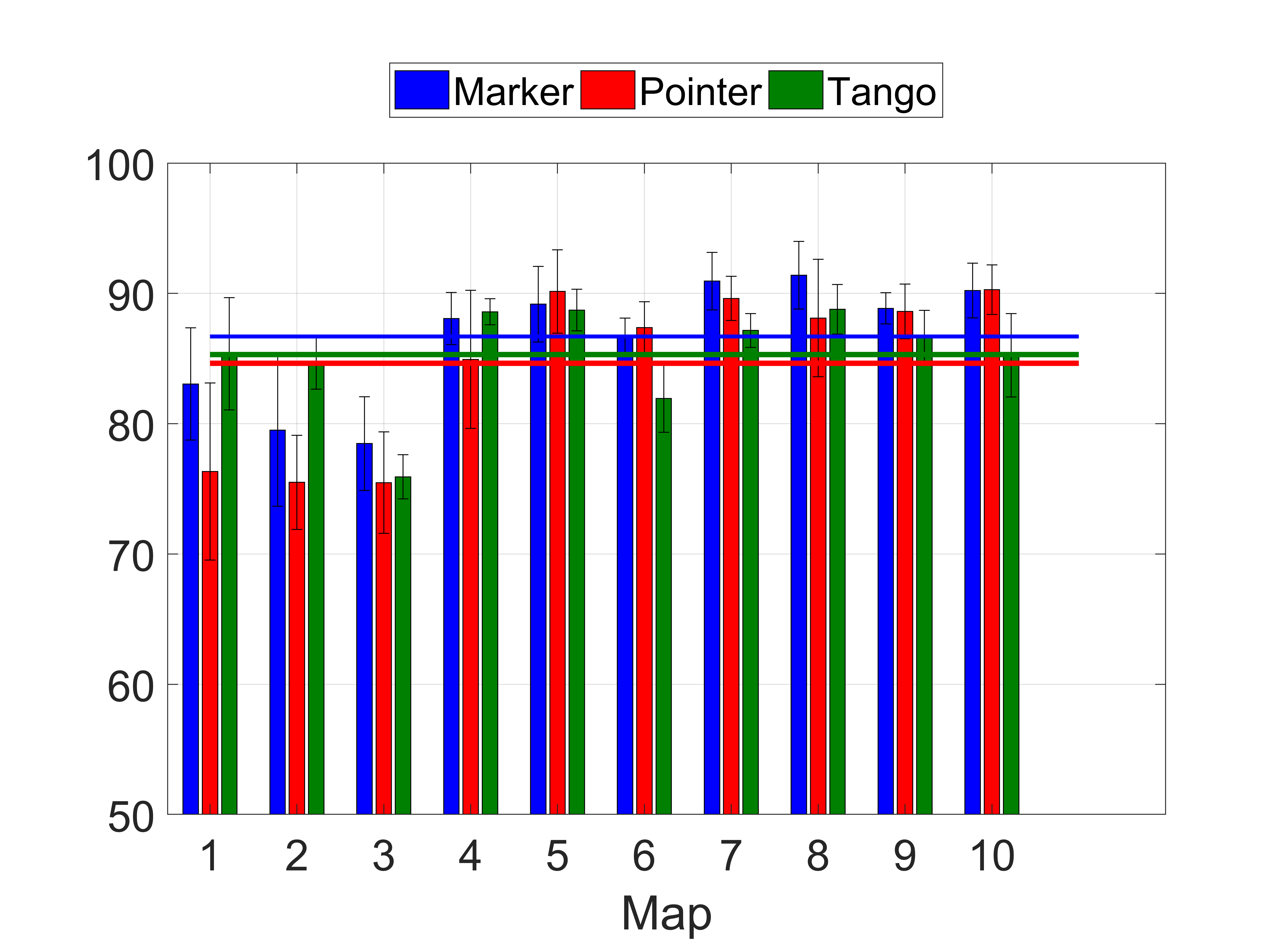}   
	\caption{Accuracy results for the proposed approach compared to the baseline approaches. Bars show the average accuracies per map of the self-recorded dataset. The horizontal lines indicate the overall averages per interaction device.}
	\label{fig:accEval}
\end{figure}

The detailed accuracy results of \textit{Experiment~2} depending on the maps are visualized in Fig.~\ref{fig:accEval}. Horizontal lines represent the overall means of the interaction devices that go along with the values in Fig.~\ref{fig:accComparison}. The mean accuracies per map do not significantly differ from their overall means. This shows that the accuracy of the teaching methods is independent of the shape and length of the virtual border. But there is one exception: the accuracies of the marker and laser pointer approaches significantly fall below their means for maps 1~-~3. These are maps with short virtual borders (4~m~-~6~m), and it is hard to accurately guide the robot on such a small area. In this case, our Tango approach is more flexible since the user directly interacts with the environment. Thus, yielding a higher accuracy for short virtual borders except of map~3 due to a localization error of the Tango device. It is also apparent that all interaction devices feature a small standard deviation for all maps ranging from 1.0\% to 6.8\%. Therefore, we conclude that the variations we introduced in the teaching process do not affect the accuracy.\par

Fig~\ref{fig:accVisualization} visualizes some exemplary accuracy results of \textit{Experiment~2} for the different interaction devices. These underline the high accuracies of all interaction methods. Inaccuracies occur due to localization errors (either mobile robot or Tango device) and the interaction of the user. The figure also reveals another percularity of the Tango approach: in this case, there is a constant rotational error for all maps. This is caused by an inaccuracy of the manual registration between the \textit{ADF} and \textit{Map} coordinate frames. If this error could be fixed in the future, even more accurate results could be possible.

\subsection{Teaching Time}
\begin{figure}
	\centering
	\includegraphics[width=0.46\textwidth]{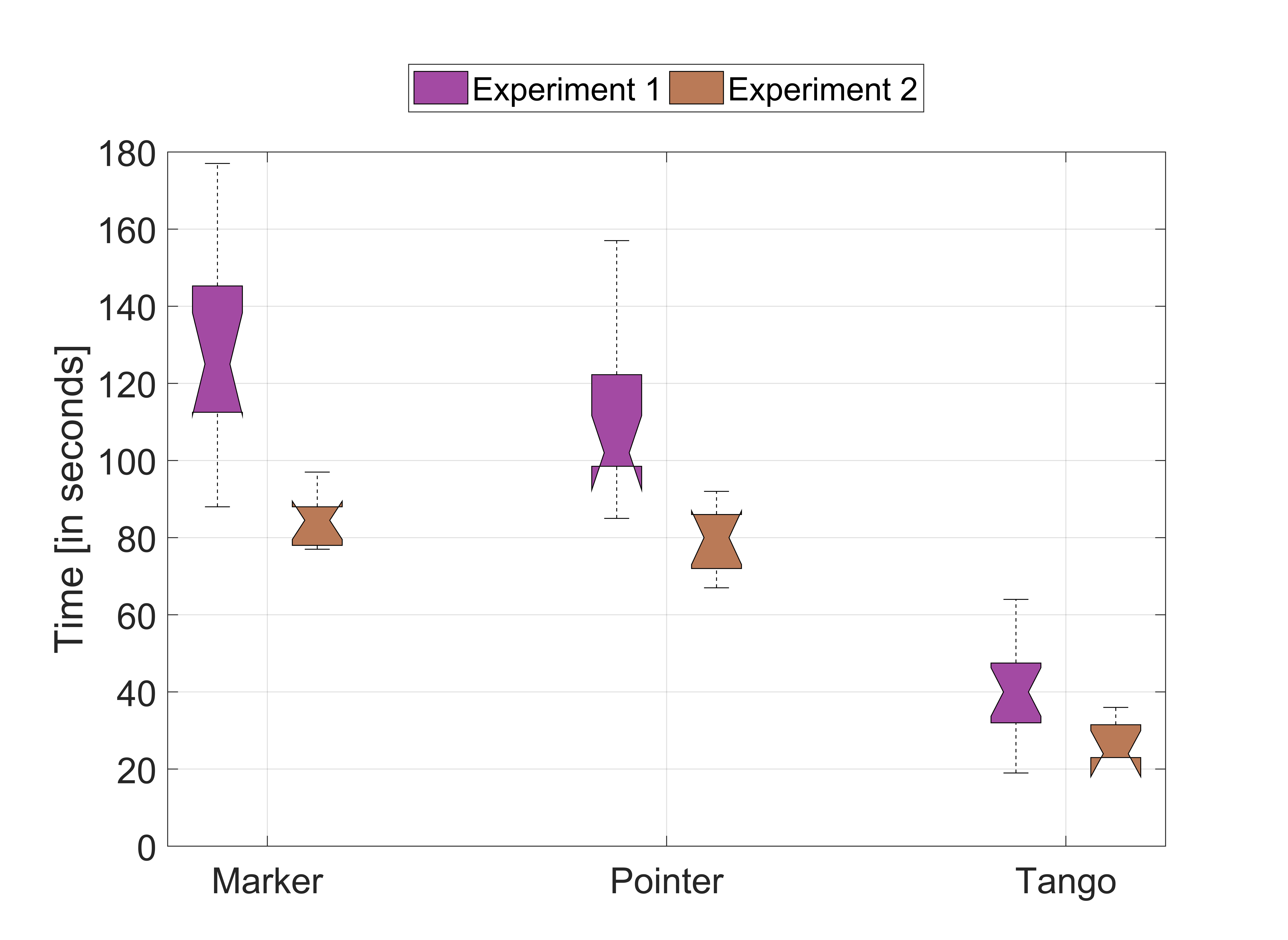}   
	\caption{Comparison of the teaching time for both experiments depending on the interaction device.}
	\label{fig:timeComparison}
\end{figure}

The second criterion is considered to answer the question of \textit{how much time does it take a user to teach virtual borders}. Fig.~\ref{fig:timeComparison} shows the teaching time for each experiment depending on the interaction device. The results of \textit{Experiment~1} are shown as purple box plots. Both baseline methods feature a high average teaching time (marker: 129~s, pointer: 112~s), while our Tango approach has an average teaching time of 40~s. Similarly, the baseline methods feature an equally high teaching time in \textit{Experiment~2} (marker: 85~s, pointer: 79~s) and the Tango approach only 27 s. These results reveal two insights: (1) our Tango approach is significantly faster compared to the baseline methods and (2) experience in handling the interaction device in the teaching process can reduce the teaching time.\par

\begin{figure}
	\centering
	\includegraphics[width=0.46\textwidth]{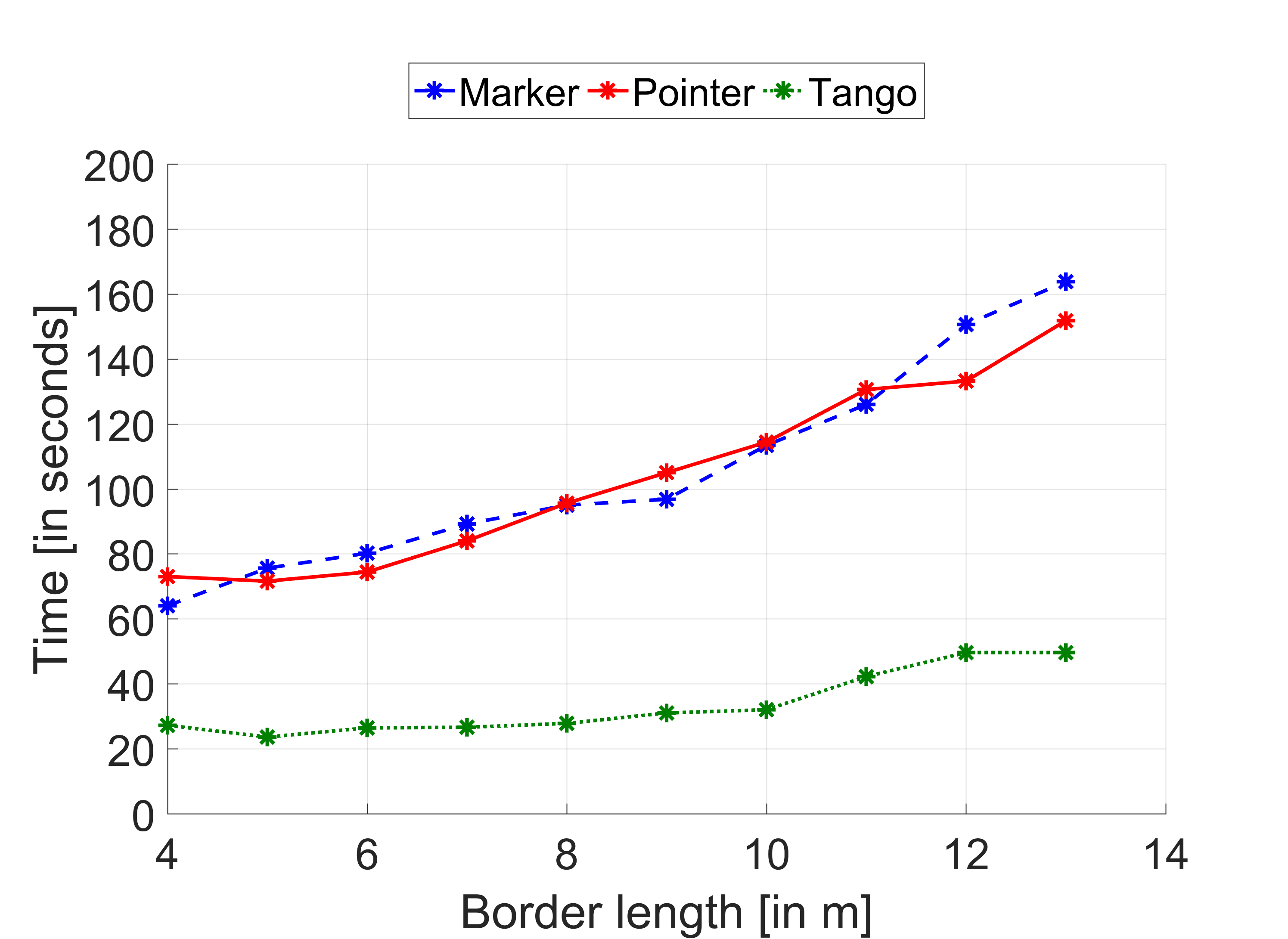}   
	\caption{Average teaching times of the interaction devices dependent on the border length.}
	\label{fig:timeEval}
\end{figure} 
Fig.~\ref{fig:timeEval} shows the detailed teaching time of \textit{Experiment~2} dependent on the border length. While there is a linear relationship between the teaching time and the border length, our approach features a smaller gradient. This is due to the nature of the baseline methods: users interact with the mobile robot to define the border points and are limited by the velocity of the robot. In contrast to the baseline methods, users directly interact with the environment using the Tango tablet making it independent of the robot's velocity. As a result, our proposed method is significantly faster than the comparative methods, e.g. it takes 50~s for teaching a 13~m long border using a Tango device, while the marker approach takes 164~s and the laser pointer approach 152~s for the same border. This corresponds to a speedup of 3.3 and 3.1, respectively. The speedup factors for the other maps with respect to the baseline approaches is presented in Tab. \ref{tab:speedup}. The proposed Tango method is 3.1 times faster on average than the marker and laser pointer approach.

% Table generated by Excel2LaTeX from sheet 'Tabelle1'
\begin{table}[htbp]
  \centering
  \caption{Speedup compared to the baseline approaches.}
    \begin{tabular}{l|p{0.2cm}p{0.2cm}p{0.2cm}p{0.2cm}p{0.2cm}p{0.2cm}p{0.2cm}p{0.2cm}p{0.2cm}p{0.2cm}p{0.2cm}}         
          & \multicolumn{10}{c}{Border length [in m]}                                            &  \\
    Baseline  & 4     & 5     & 6     & 7     & 8     & 9     & 10    & 11    & 12    & 13    & \multicolumn{1}{l}{Avg.} \\
    \toprule
    Marker & 2.4   & 3.2   & 3.0   & 3.4   & 3.4   & 3.1   & 3.5   & 3.0   & 3.0   & 3.3   & \textbf{3.1} \\
    Pointer & 2.7   & 3.0   & 2.8   & 3.2   & 3.4   & 3.4   & 3.6   & 3.1   & 2.7   & 3.1   & \textbf{3.1} \\
    \end{tabular}%
  \label{tab:speedup}%
\end{table}%

Finally, all interaction devices feature a mean standard deviation for all maps ranging from 4~s (Tango) to 8~s (marker). Thus, the variations during the runs do not affect the teaching time significantly.

\subsection{Correctness}
\begin{figure*}
\centering
		\centering
        \begin{subfigure}[b]{0.32\textwidth}
                \centering
                \includegraphics[width=\textwidth]{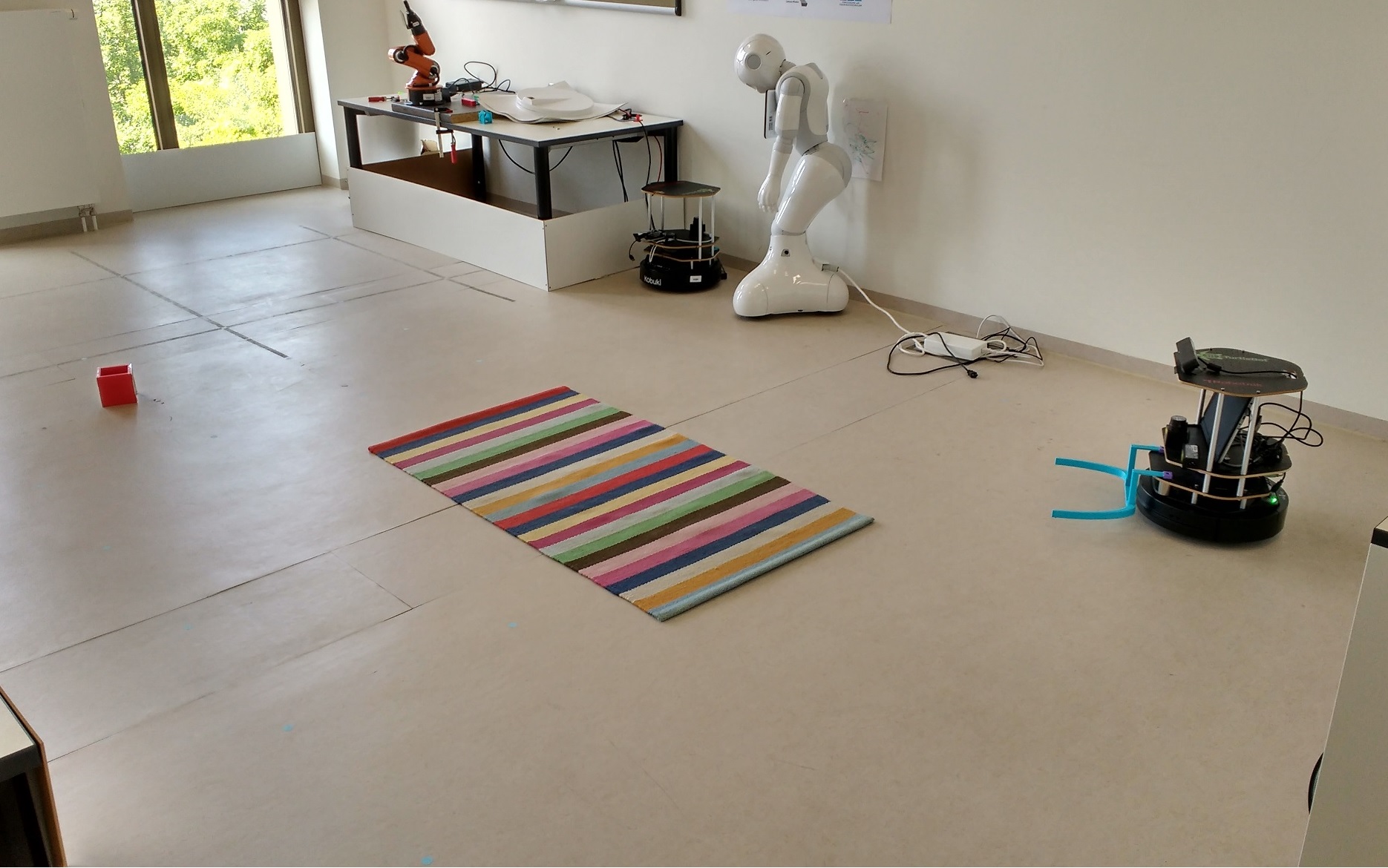}
                \caption{Lab environment} 
                \label{fig:lab}               
        \end{subfigure}   
        \begin{subfigure}[b]{0.32\textwidth}
                \centering
                \includegraphics[width=\textwidth]{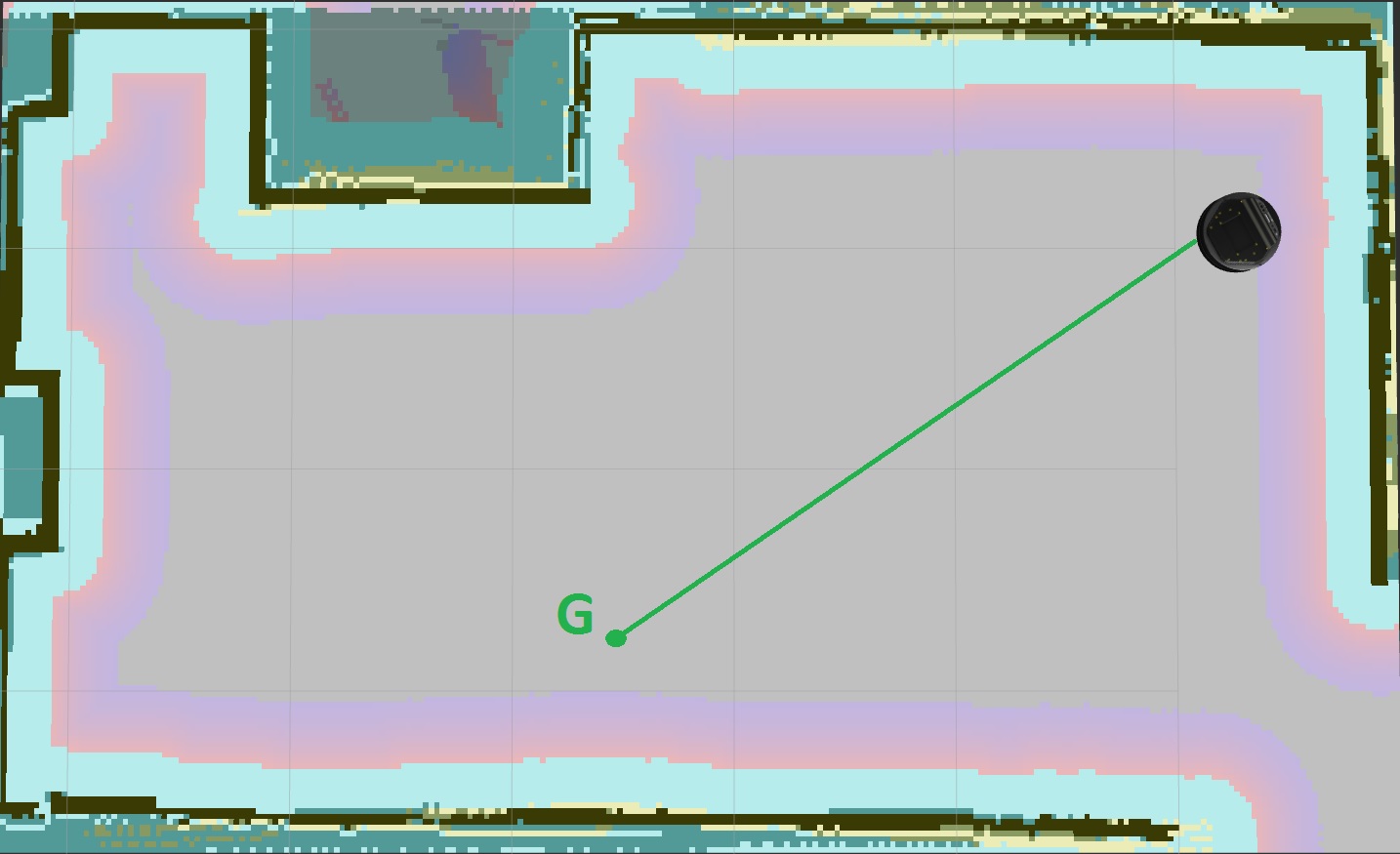}
                \caption{Costmap before teaching process}   
                \label{fig:beforeCostmap}                             
        \end{subfigure} 
		\centering
        \begin{subfigure}[b]{0.32\textwidth}
                \centering
                \includegraphics[width=\textwidth]{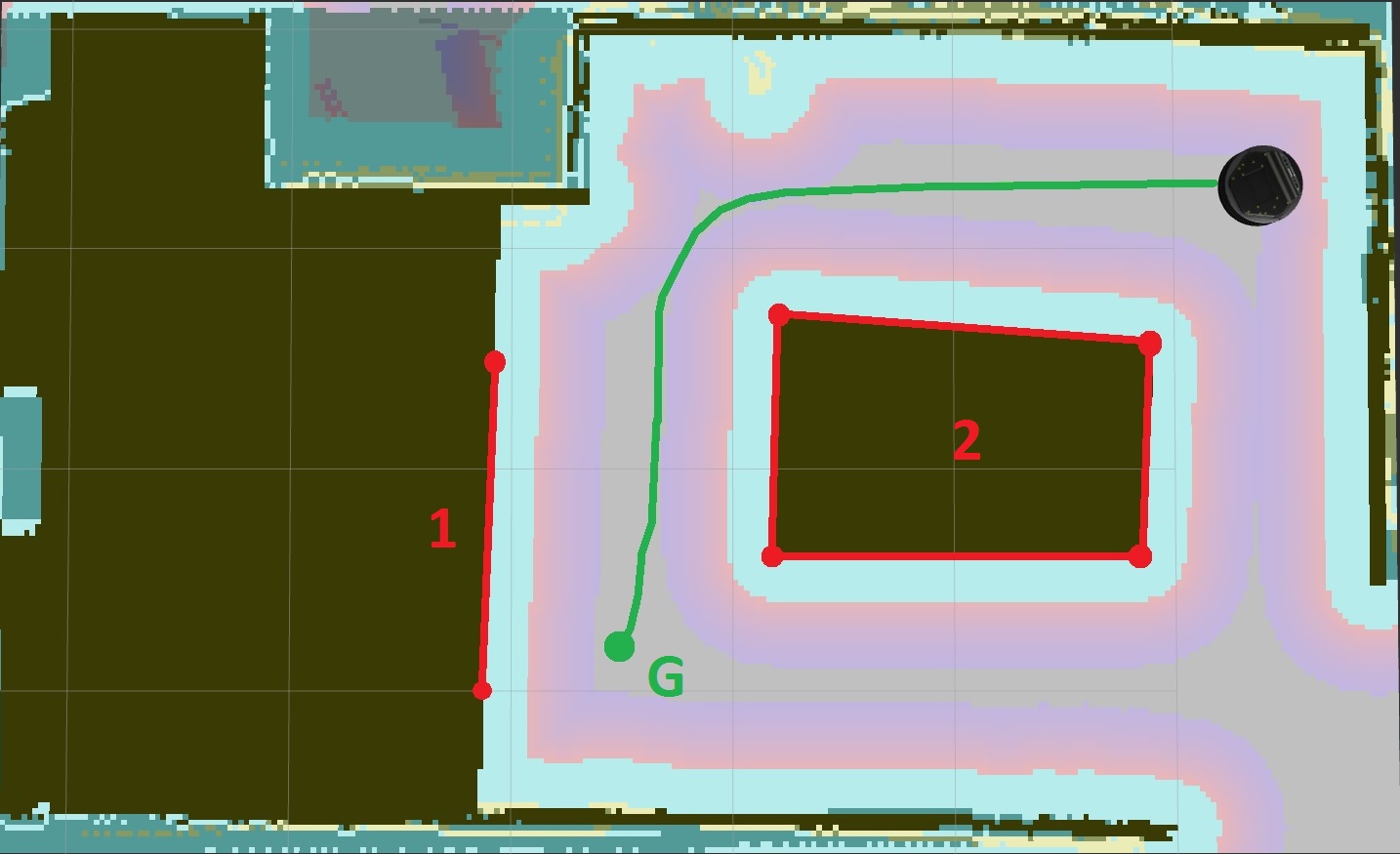}
                \caption{Costmap after teaching process}
                \label{fig:afterCostmap}                
        \end{subfigure}                             
        \caption{Lab environment and costmaps for a navigation scenario. (a) shows the lab environment with a mobile robot, a carpet and a navigation goal (red cube). (b) and (c) visualize the global costmaps for the same navigation goal before and after teaching virtual borders. Two virtual borders, a separating curve (1) and a polygon (2), shown as red lines have been integrated by the user during the teaching process. The navigation paths are drawn as a green lines.} 
        \label{fig:costmap}
\end{figure*}
The correctness is evaluated to answer the question \textit{whether the teaching process successfully changes the navigational behavior of the mobile robot}. This depends on the resulting posterior map of the teaching process. Therefore, we set up a simple navigation scenario as shown in Fig.~\ref{fig:costmap} where the mobile robot is instructed to navigate to the red cube in the left image of the lab environment. The centered image shows the global costmap and the path to the navigation goal based on the physical OGM $M_{prior}$ of the environment. As expected, the mobile robot crosses the carpet area while driving to its goal because it is the shortest path (the path with the fewest costs). In order to avoid the robot from crossing the carpet, we use the posterior OGM $M_{posterior}$ from the teaching process visualized in Fig.~\ref{fig:teaching} as basis for a global costmap. This costmap and the calculated path to the same goal is shown in the right image of Fig.~\ref{fig:costmap}. The mobile robot now circumvents the carpet as desired. The results show that the teaching method successfully integrates the virtual borders into the global OGM and changes the navigational behavior of the mobile robot. Thus, a user can easily control the workspace of a mobile robot. Note that the actual teaching process is independent of a concrete path planner for navigation.

%\addtolength{\textheight}{-1.5cm}   % This command serves to balance the column lengths

\section{Discussion, Conclusions \& Future Work}
We developed a teaching method for incorporating virtual borders into given OGMs using a RGB-D device in combination with an AR application. This allows non-expert users to flexibly and interactively define arbitrary virtual borders in their mobile robots' workspaces. Thus, users can prevent their robots to enter certain places, e.g. bath rooms or carpet areas, which gives them the ability to effectively control their mobile robots in a simple way allowing human-aware navigation in human-centered environments. We compared our method with other approaches, and the results revealed an accuracy on the same level as the baseline methods while featuring a significantly lower teaching time. This also holds for different variations in the teaching process and different border lengths and shapes. Furthermore, there is evidence that experience in handling the teaching method increases the accuracy and reduces the teaching time. Finally, our method integrates a visual feedback system based on an AR application and does not rely on additional equipment for teaching, e.g. cameras in the environment or on a robot.\par

A weakness of the proposed method is the requirement concerning the registration between the \textit{Map} and \textit{ADF} coordinate frames (see Sect.~\ref{sec:requirements}). Although this is an initial step that needs to be performed only once, this is not suitable for a non-expert and thus needs to be addressed in the future. It would be interesting to develop an automatic registration method to circumvent this limitation. Besides, our method requires special hardware, i.e. a Tango-enabled device, limiting the potential number of users. Nonetheless, major companies currently release AR toolkits (ARCore by Google and ARKit by Apple) working without specialized hardware. Thus, our proposed method could be widely deployed on common smartphones and tablets without additional costs for the user. Our future work also focuses on a more comprehensive evaluation of the user's perspective with respect to the different interaction devices, e.g. usability aspects.
 
%%%%%%%%%%%%%%%%%%%%%%%%%%%%%%%%%%%%%%%%%%%%%%%%%%%%%%%%%%%%%%%%%%%%%%%%%%%%%%%%
%\section*{ACKNOWLEDGMENT}

%%%%%%%%%%%%%%%%%%%%%%%%%%%%%%%%%%%%%%%%%%%%%%%%%%%%%%%%%%%%%%%%%%%%%%%%%%%%%%%%
\bibliography{bibo}   
\bibliographystyle{IEEEtran}

\end{document}